\documentclass[runningheads]{llncs}
\usepackage[T1]{fontenc}
\usepackage{graphicx}
\usepackage{amsmath}
\usepackage{amsfonts} 
\usepackage{multirow}
\usepackage{booktabs}
\usepackage{adjustbox}
\usepackage{multicol}
\usepackage{xcolor}
\usepackage{comment}
\usepackage{url}
\usepackage[misc,geometry]{ifsym}
\usepackage{hyperref}

\newcommand{\joinR}{\hspace{-.1em}}
\newcommand{\RomanI}{I}
\newcommand{\RomanII}{\mbox{\RomanI\joinR\RomanI}}

\begin{document}

\title{DEGNN: Dual Experts Graph Neural Network Handling Both Edge and Node Feature Noise}
%
\titlerunning{Dual Experts GNN Handling Both Edge and Node Feature Noise}
%
\author{anonymous authors}
\author{Tai Hasegawa\inst{1,2}\orcidID{0000-0003-4089-0186} \and
Sukwon Yun\inst{3}\orcidID{0000-0002-5186-6563} \and
Xin Liu \inst{2}\orcidID{0000-0002-2336-7409}\Letter \and
Yin Jun Phua\inst{1}\orcidID{0000-0003-1178-8238} \and
Tsuyoshi Murata\inst{1,2}\orcidID{0000-0002-3818-7830}
}

\authorrunning{T. Hasegawa et al.}

\institute{Department of Computer Science, Tokyo Institute of Technology, Japan\\
\email{\{hasegawa.t@net,phua@,murata@\}.c.titech.ac.jp}\\
\and
Artificial Intelligence Research Center, AIST, Japan\\
\Letter\email{xin.liu@aist.go.jp}\\
\and
Industrial and Systems Engineering, KAIST, Republic of Korea\\
\email{swyun@kaist.ac.kr}}

%

\maketitle              
\begin{abstract}
Graph Neural Networks (GNNs) have achieved notable success in various applications over graph data. However, recent research has revealed that real-world graphs often contain noise, and GNNs are susceptible to noise in the graph. To address this issue, several Graph Structure Learning (GSL) models have been introduced. While GSL models are tailored to enhance robustness against edge noise through edge reconstruction, a significant limitation surfaces: their high reliance on node features. This inherent dependence amplifies their susceptibility to noise within node features. Recognizing this vulnerability, we present DEGNN, a novel GNN model designed to adeptly mitigate noise in both edges and node features. The core idea of DEGNN is to design two separate experts: an edge expert and a node feature expert. These experts utilize self-supervised learning techniques to produce modified edges and node features. Leveraging these modified representations, DEGNN subsequently addresses downstream tasks, ensuring robustness against noise present in both edges and node features of real-world graphs. Notably, the modification process can be trained end-to-end, empowering DEGNN to adjust dynamically and achieves optimal edge and node representations for specific tasks. Comprehensive experiments demonstrate DEGNN's efficacy in managing noise, both in original real-world graphs and in graphs with synthetic noise. 


\end{abstract}

\section{Introduction}
\label{sec:introduction}
Graphs are essential data structures for modeling a wide range of real-world phenomena, such as social networks, transportation networks, and chemical molecules. Graph Neural Networks (GNNs) have emerged as a powerful paradigm for modeling such graphs, primarily due to their message-passing mechanism that aggregates node representations via edges.
These GNNs can be applied to various tasks, including node classification~\cite{kipf2017semi,maurya2022simplifying}, link prediction~\cite{zhang2018link}, node ranking~\cite{maurya2019fast,maurya2021graph}, community detection~\cite{choong2018learning}, and graph classification~\cite{zhang2018end}.

Despite their success, it is well known that the quality of real-world graph data is often unreliable~\cite{marsden1990network}. In other words, real-world graphs are known to contain noise.
For instance, in citation networks, references to unrelated papers can introduce noise in the form of inaccurate edges.
Recent studies on adversarial attacks and defenses have highlighted the susceptibility of GNNs to noise within graphs~\cite{dai2018adversarial,jin2021adversarial}.
In response, Graph Structure Learning (GSL) has been developed as a method to optimize graph structure, thus improving graph representations and ensuring more resilient predictions amidst edge noise.

\begin{figure}[!t]
\centering
\includegraphics[width=0.7\textwidth]{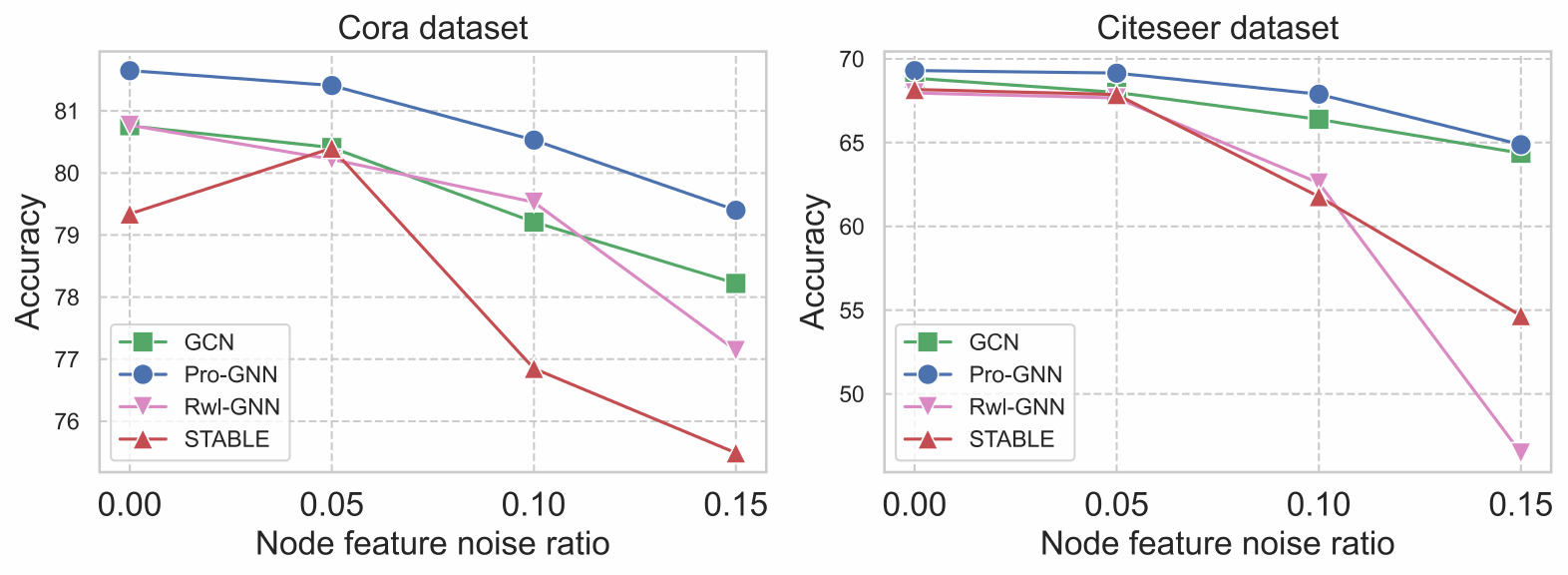}
\vspace*{-3mm}
\caption{A comparison of semi-supervised node classification between GCN and GSL models on Cora and Citeseer dataset when there is noise in the node features.}
\label{fig:nf_noise}
\end{figure}

Besides edge noise, node features can also contain noise. For instance, in social networks, users might provide inconsistent, overstated, or even false information about their interests or attributes, which introduces noise into the node features. GSL models often struggle with such node feature noises. The vulnerability of GSL models against node feature noise arises as these models disseminate noisy node features via message passing scheme. Additionally, their reliance on node features to rewire edges compounds the challenge.
As illustrated in Figure~\ref{fig:nf_noise}
(refer to Section~\ref{sec:experiments} for the experiment details), recent GSL models such as Pro-GNN~\cite{jin2020graph}, Rwl-GNN~\cite{runwal2022robust}, and STABLE~\cite{li2022reliable} tend to underperform in prediction accuracy when compared to traditional GNN model, GCN~\cite{kipf2017semi} when the node feature noise increases. Although Pro-GNN marginally surpasses the performance of GCN, there remains ample scope for improvements.

In this paper, we introduce Dual Experts Graph Neural Network (DEGNN)\footnote{Codes are available at: \url{https://github.com/TaiHasegawa/DEGNN}}, a novel GNN model crafted to offer robust predictions irrespective of the presence or absence of noise in edges, nodes, or both. At the heart of DEGNN is its distinctive architecture, which employs specialized branches, termed "experts", to individually learn and refine node features and edges. Using a self-supervised learning approach, these experts are seamlessly integrated and co-trained end-to-end, ensuring task-specific optimization.
Our contributions are summarized as follows:
\begin{itemize}
    \item We highlight the susceptibility of GSL models to node feature noise based on preliminary experiments.
    \item We introduce DEGNN, a novel GNN that offers robust predictions irrespective of the presence or absence of noise in nodes, edges, or both, by individually addressing these noises through a self-supervised learning approach. 
    \item Through comprehensive experiments on real-world datasets, we establish that DEGNN consistently delivers stable predictions, outperforming state-of-the-art models in the presence of either type of noise.
\end{itemize}
\section{Related Work}
\subsection{Graph Neural Networks}
GNNs have emerged as a powerful tool for learning from graph-structured data, effectively capturing the complex relationships and interdependencies between nodes ~\cite{zhou2020graph}. Their successful development across numerous practical fields underscores their extensive applicability and effectiveness~\cite{marcheggiani2017encoding,rakhimberdina2020population,djenouri2023hybrid,jin2021predicting,fan2019graph}. Generally, GNNs can be classified into two categories: spectral-based methods~\cite{defferrard2016convolutional,kipf2017semi} and spatial-based methods~\cite{hamilton2017inductive,veličković2018graph}. Spectral-based GNNs hinge upon spectral graph theory~\cite{chung1997spectral} and employ spectral convolutional neural networks. To streamline the intricacies of spectral-based GNNs, various techniques have emerged, including ChebNet~\cite{defferrard2016convolutional} and GCN~\cite{kipf2017semi}.



On the other hand, spatial-based GNNs are engineered to tackle challenges related to efficiency, generality, and flexibility, as highlighted in~\cite{wu2020comprehensive}.
They achieve graph convolution operation through neighborhood aggregation.
For instance, GraphSAGE~\cite{hamilton2017inductive} selectively samples a subset of neighbors to grasp local features, while GAT~\cite{veličković2018graph} employs an attention mechanism for adaptive neighbor aggregation.
However, these models are susceptible to edge noise.

\subsection{Graph Structure Learning}
The purpose of GSL is to improve prediction accuracy by modifying the given graph into an optimal structure.
In this paper, we primarily focus on GNN-based graph structure learning models.
LDS~\cite{Franceschi2019LearningDS} jointly optimizes the probability for each node pair and the parameters of GNNs in a bilevel way.
Pro-GNN~\cite{jin2020graph} aims to learn the optimal graph structure by incorporating several regularizations, such as low-rank sparsity and feature smoothness.
Gaug-M~\cite{zhao2021data} directly computes the edge weights by taking the inner product of node embeddings.
STABLE~\cite{li2022reliable} utilizes self-supervised learning to acquire node embeddings and then modifies the graph structure based on their similarities. These obtained node embeddings and the modified graph structure are employed in downstream tasks.
Each of these models demonstrates the ability to robustly predict against edge noise, as shown in their respective papers. However, as elucidated in Section~\ref{sec:introduction}, their pronounced reliance on node features during edge rewiring inherently exposes them to vulnerabilities in situations characterized by noisy node features.



\section{The Proposed Model}
In this section, we introduce our proposed approach, DEGNN. We begin by formulating the problem definition, followed by an overview of the model, and then provide a detailed description of its architecture and learning procedure.

\subsection{Problem Definition}
Let $\mathcal{G}=\{\mathcal{V}, \mathcal{E}, X\}$ represent an undirected graph, where $\mathcal{V}=\{v_i\}^N_{i=1}$ is the set of N nodes, $\mathcal{E}$ is the set of edges, $X = [x_1, ..., x_N] \in \mathbb{R}^{N \times D}$ indicates the node feature matrix and each $x_i \in \mathbb{R}^{D}$ is the feature vector of node $v_i$.
The set of edges is represented by an adjacency matrix $A \in \{0,1\}^{N \times N}$, where $A_{ij}$ denotes the connection between nodes $v_i$ and $v_j$.
Following the common semi-supervised node classification setting, only a small portion of nodes $\mathcal{V}_L = \{v_i\}^l_{i=1}$ are associated with the corresponding labels $\mathcal{Y}_L = \{y_i\}^l_{i=1}$ while the rest of the nodes $\mathcal{V}_U = \{v_i\}^{N}_{i=l+1}$ are unlabeled.

Given graph $\mathcal{G}=\{\mathcal{V}, \mathcal{E}, X\}$ and the available labels $\mathcal{Y}_L$, the goal of graph structure learning, aimed at refining node features and graph structure, is to learn optimal node embeddings $H \in \mathbb{R}^{N \times D'}$ with hidden dimension $D'$, a modified adjacency matrix $S \in \mathbb{R}^{N \times N}$, and the GNN parameters $\theta$ in order to improve the predictive accuracy of $\hat{\mathcal{Y}}_L$.
The objective function can be formulated as
\begin{equation}
    \min_{\theta, S, H}\mathcal{L}(A, X, \mathcal{Y}_L) = \sum_{v_i \in V_L}\ell(f_{\theta}(H, S)_i, y_i),
\end{equation}
where $f_{\theta}: \mathcal{V}_L \rightarrow \mathcal{Y}_L$ is a function learned by downstream GNNs that maps nodes to the set of labels, $f_{\theta}(H, S)_i$ is the prediction of node $v_i$, and $\ell$ is the loss measuring difference between prediction and true label such as cross-entropy. 
\subsection{Overview}
\begin{figure}[!t]
\centering
\includegraphics[width=\textwidth]{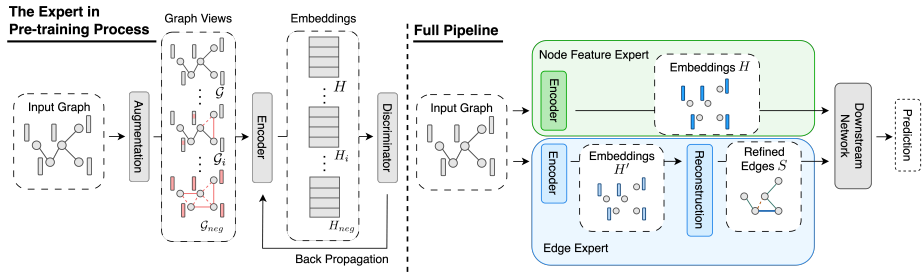}
\vspace*{-3mm}
\caption{The overview of DEGNN and the expert in its pre-training process.}
\label{fig:model-overview}
\end{figure}
The overview of DEGNN and the expert in its pre-training process are illustrated in Figure~\ref{fig:model-overview}.
The input graph is passed through both the node feature expert and the edge expert. The node feature expert outputs node embeddings $H$, while the edge expert produces the modified adjacency matrix $S$.
Using the obtained $H$ and $S$, the downstream network makes predictions for a specific task. 
Traditionally, models have primarily leaned on learned node embeddings for edge reconstruction~\cite{li2022reliable}. However, when these embeddings are derived from noisy node features, the resulting in reconstructed edges often produce sub-optimal outcomes, potentially that are unintended dependencies.
Our proposed dual experts design aims to eliminate these dependencies and learn both node representations and edges to be optimal.
These two experts can be trained with the downstream network end-to-end, allowing each of them to acquire representations suitable for the specific task.
The details of the model are described in the following sections.

\subsection{Node Feature Expert}
A straightforward approach to predict robustly against node feature noise is to generate node embeddings $H$ that capture node features more effectively than the input node features $X$.
To obtain the node embeddings $H$, we utilize self-supervised learning, since it enables the model to achieve better performance, generalization, and robustness across a wide range of downstream tasks~\cite{liu2022graph}.

\subsubsection{Graph Augmentation}
Generating views is a key component of self-supervised learning methods.
For instance, in the field of computer vision~\cite{berthelot2019mixmatch}, views are created by rotating or cropping images. This allows for mitigating the impact of differences in angles and scales on the model's predictions.
Similarly, for graph data, we presume that by generating views and exposing the model to modified edges and nodes, it can achieve enhanced generalizability, equipping it with a stronger capacity to manage noise emanating from both edges and nodes during predictions.
Therefore, we generate graphs of positive view $\mathcal{G}_1 = (\tilde{A}, X)$ with noise added to the edges, $\mathcal{G}_2 = (A, \tilde{X})$ with noise added to the node features, and $\mathcal{G}_3 = (\tilde{A}, \tilde{X})$ with noise added to both edges and node features, where $\tilde{A} \in \{0,1\}^{N \times N}$ and $\tilde{X} \in \mathbb{R}^{N \times D}$ denote the noisy adjacency matrix and noisy node features, respectively.

Our proposed model augments edges by randomly rewiring them.
Formally, we first create a mask matrix $P \in \{0,1\}^{N \times N}$ to rewire edges, where $P$ is obtained from a Bernoulli distribution $P_{ij} \sim \mathcal{B}(p)$ with a hyper-parameter $p$ that controls the rewiring probability.
And then, $\tilde{A}$ is calculated as follows:
\begin{equation}
\tilde{A} = (1 - A) \odot P + A \odot (1 - P),
\end{equation}
where $\odot$ is the Hadamard product.
For node feature augmentation, unlike other studies that use node feature masking~\cite{zhu2020deep,you2020graph},
in this paper, we shuffle elements in each row of $X$ randomly with the probability $q$ to replicate scenarios where there is noise in the node features.

Besides positive graphs, we generate a negative graph $\mathcal{G}_{neg} = (A_{neg}, X_{neg})$ that is entirely different from the original graph $\mathcal{G}$, where $A_{neg} \in \{0,1\}^{N \times N}$ and $X_{neg} \in \mathbb{R}^{N \times D}$ denote the negative adjacency matrix and negative node features, respectively. $A_{neg}$ is given by $A_{neg} = (1 - A) \odot P_{neg}$,
where $P_{neg} \in \{0,1\}^{N \times N}$ is a mask matrix that is obtained from a Bernoulli distribution $P_{negij} \sim \mathcal{B}(\frac{\lVert \mathcal{E} \rVert}{N^2-\lVert \mathcal{E} \rVert})$. $X_{neg}$ is obtained through row-wise shuffling of $X$. 

\subsubsection{Encoder}
The encoder $f_{\phi}$ parameterized by $\phi$ learns embeddings for each of the generated views.
In this paper, we use a 1-layer GCN~\cite{kipf2017semi} as the encoder, which is formulated as follows:
\begin{equation}
f_{\phi}(X, A) = \sigma(\hat{D}^{-1/2}\hat{A}\hat{D}^{-1/2}XW),
\end{equation}
where $\hat{A} = A + I_N$, $\hat{D} = D + I_N$, D is the degree matrix of $A$, $I_N$ is the identity matrix, $\sigma$ is a non-linear activation function and $W$ is weight matrix transforming raw features to embeddings.
The embeddings for the original graph and each view $H, H_1, H_2, H_3, H_{neg} \in \mathbb{R}^{N \times D'} $ are obtained using the encoder as follows:
\begin{multicols}{2}
\begin{equation}
H = f_{\phi}(X, A)
\end{equation}
\begin{equation}
H_1 = f_{\phi}(X, \tilde{A})
\end{equation}
\begin{equation}
H_2 = f_{\phi}(\tilde{X}, A)
\end{equation}
\begin{equation}
H_3 = f_{\phi}(\tilde{X}, \tilde{A})
\end{equation}
\begin{equation}
H_{neg} = f_{\phi}(X_{neg}, A_{neg})
\end{equation}
\end{multicols}


\subsubsection{Objective Function}
Following other graph contrastive learning methods~\cite{suresh2021adversarial,li2022reliable}, we train the encoder to maximize the mutual information between the original graph $\mathcal{G}$ and positive graphs $\mathcal{G}_1$, $\mathcal{G}_2$ and $\mathcal{G}_3$ while minimizing agreement between original graph $\mathcal{G}$ and negative graph $\mathcal{G}_{neg}$. Formally, the objective function for node feature expert can be formulated via using binary cross-entropy loss between positive samples and negative samples as follows:
\begin{equation}
\mathcal{L}_N = -\frac{1}{2N}\sum_{i=1}^{N}(\frac{1}{3}\sum_{j=1}^{3}(log\mathcal{D}(h_{i}, h_{i}^{j}) + log(1 - \mathcal{D}(h_{i}, h_{i}^{neg})))),
\end{equation}
where $h_i$, $h_i^j$, $h_i^{neg}$ are the embeddings of node $v_i$ in $\mathcal{G}$, $\mathcal{G}_j$ and $\mathcal{G}_{neg}$, and $\mathcal{D}$ is the discriminator built upon an inner product, i.e., $\mathcal{D}(a, b)=ab^T$. In conclusion, by training the node feature expert, $f_{\phi}$, we derive a noise-robust node embedding $H$ that subsequently serves as input for the downstream network.

\subsection{Edge Expert}
To differentiate between the impacts of node feature noise and edge noise, and to adeptly address each situation, we further implement an edge expert. The edge expert learns node embeddings $H'$ using self-supervised learning as with the node feature expert. More precisely, $H'$ is obtained through an encoder $f_{\psi}$ with different parameters from the node feature expert $f_{\phi}$, and the objective function for edge expert using the binary cross-entropy loss $L_E$ can be formulated as follows:
\begin{equation}
\mathcal{L}_E = -\frac{1}{2N}\sum_{i=1}^{N}(\frac{1}{3}\sum_{j=1}^{3}(log\mathcal{D}(h'_{i}, h_{i}^{j'}) + log(1 - \mathcal{D}(h'_{i}, h_{i}^{neg'})))),
\end{equation}
where $h'_i$, $h_i^{j'}$, $h_i^{neg'}$ are the embeddings of node $v_i$ in $\mathcal{G}$, $\mathcal{G}_j$ and $\mathcal{G}_{neg}$ obtained with encoder $f_{\psi}$.

\subsubsection{Reconstruction}
Once the high-quality embedding $H'$ from the edge expert is obtained, it is used to reconstruct the graph structure.
This reconstruction process rewires the edges using the pairwise similarity in the embeddings $H'$ under the homophily assumption~\cite{mcpherson2001birds}, which posits that nodes with similar features are more likely to be connected.

In the beginning, we compute the cosine similarity matrix $B \in \mathbb{R}^{N \times N}$, where $B_{ij}$ represents the cosine similarity between $H_i'$ and $H_j'$.
Next, we remove the $k\%$ ($k$ is a hyper-parameter) of the edges with the smallest cosine similarity between pairs of nodes from the original edge set $\mathcal{E}$, resulting in a sparser adjacency matrix $\tilde{S} \in \{0,1\}^{N \times N}$ as follows:
\begin{equation}
\tilde{S} = A \odot M_1,
\end{equation}
where $M_1 \in \{0,1\}^{N \times N}$ represents the mask matrix for edge deletion, where $M_{1ij}$ is 1 if its cosine similarity ranks over the smallest $k$\%, and 0 otherwise.

Finally, to obtain the modified adjacency matrix $S$, we add the same number of edges that were removed i.e., $k * |\mathcal{E}|$ with the highest cosine similarity between pairs of nodes from the set of node pairs $\mathcal{E}' = V \times V \setminus \mathcal{E}$ that are not included in the edge set $\mathcal{E}$.
Formally, $S$ is obtained as follows:
\begin{equation}
S = \tilde{S} + B \odot M_2,
\end{equation}
where $M_2 \in \{0,1\}^{N \times N}$ is the mask matrix for edge addition, where $M_{2ij}$ is 1 for the edges within top cosine similarity ($k * |\mathcal{E}|$), and 0 otherwise.
Here, it is important that the encoder in the edge expert can be trained through backpropagation from the downstream network, so the reconstruction needs to be differentiable. Consequently, the value associated with newly introduced edges in $S$ is not clamped and retains its cosine similarity. To sum up, through training edge expert, $f_{\psi}$, we obtain a modified adjacency matrix, $S$, which is then utilized as input for the downstream network.

\subsection{Downstream Network}
Once we obtain the node embeddings $H$ and the modified adjacency matrix $S$, we now use them as input for the downstream network (i.e., GNNs). It is worth noting that our proposed method allows the use of any GNNs such as GCN~\cite{kipf2017semi} or GAT~\cite{veličković2018graph}, and also can be applied to various tasks beyond node classification, including link prediction~\cite{zhang2018link} and graph classification~\cite{zhang2018end}.
In this paper, we use a 2-layer GCN $f_{\theta}$ as the downstream network.
To tackle node classification, the model is trained to minimize the cross-entropy loss:
\begin{equation}
    \mathcal{L}_{GNN} = \sum_{v_i \in V_L}\ell(f_{\theta}(H, S)_i, y_i),
\end{equation}
where $\ell(f_{\theta}(H, S)_i, y_i)$ is the cross-entropy between the prediction and the ground-truth label for node $v_i$.

\subsection{Training Methodology}
\label{sec:training_methodology}
In this paper, we propose two variants of DEGNN with different training methods: (i) the pre-training and fine-tuning model (referred to as DEGNN-{\RomanI}), and (ii) the modular learning model (referred to as DEGNN-{\RomanII}).
DEGNN-{\RomanI}  first pre-trains the node feature expert and edge expert separately. Then, all components are jointly fine-tuned in an end-to-end manner. 
This is to enable each expert to obtain the optimal node embeddings and graph structure for downstream tasks.
During the fine-tuning process, it is trained to minimize the following objective function:
\begin{equation}
    \mathcal{L} = \mathcal{L}_{GNN} + \alpha \mathcal{L}_N + \beta \mathcal{L}_E,
\end{equation}
where $\alpha$ and $\beta$ are hyper-parameters to balance the contributions of node embedding generation and edge reconstruction, respectively.

In contrast, DEGNN-{\RomanII}  follows a two-step approach: initially training the node feature expert and edge expert and subsequently freezing them during the training of the downstream network.
This methodology empowers each expert to attain task-agnostic representations. This allows robust predictions in contexts with limited labels or large biases, given that the approach does not depend on the provided label during training—essentially, a self-supervised paradigm.


\section{Experiments}
\label{sec:experiments}
In this section, we evaluate our proposed method, DEGNN, on a variety of noisy graphs in the context of the semi-supervised node classification task.
\subsection{Experimental Setup}
\subsubsection{Datasets}
We used four open datasets, including two citation networks (i.e., Cora~\cite{zhao2021data}, Citeseer~\cite{zhao2021data}) and two co-purchasing networks (i.e., Photo~\cite{shchur2018pitfalls}, Computer~\cite{shchur2018pitfalls})
Regarding the train/validation/test split, we prepared 20 training labels for each class in all datasets. For validation and test, we used 500 and 1000 nodes, respectively.

\subsubsection{Noisy Graphs}
\label{sec:noise}
To assess the robustness of our model across various graph settings with noise, we compared models on graphs that included the following types of noise:
\begin{itemize}
    \item \textbf{Clean Graphs}: The original graphs of the datasets which may contain inherent node feature  noise and edge noise.
    \item \textbf{Edge Noisy Graphs}: We randomly remove a certain number of edges and insert the same number of fake edges.
    \item \textbf{Node Feature Noisy Graphs}: As in \cite{wu2020graph}, we added independent Gaussian noise to the node features.
    Specifically, we obtained the reference amplitude $r$ by calculating the mean of the maximum value across each node's features. For each feature dimension of each node, we introduced independent Gaussian noise $\lambda \cdot r \cdot \epsilon$, where $\epsilon \sim N(0, 1)$, and $\lambda$ represents the feature noise ratio.
    \item \textbf{Edge and Node Feature Noisy Graphs}: We introduced both the edge noise and node feature noise described above.
\end{itemize}
When adding these noises, we employed a poisoning attack, which initially prepares a graph with noise added, and used it for both training and evaluation.

\subsubsection{Baselines}
We compare the proposed DEGNN-{\RomanI} and DEGNN-{\RomanII}  with two categories of baselines: classical GNN models (i.e., GCN~\cite{kipf2017semi}, GAT~\cite{veličković2018graph} and RGCN~\cite{zhu2019robust}) and graph structure learning methods (i.e., Pro-GNN~\cite{jin2020graph}, Rwl-GNN~\cite{runwal2022robust} and STABLE~\cite{li2022reliable}).
\subsubsection{Implementation Details}
All hyper-parameters are tuned on the clean graph.
All models are trained using Adam optimizer with a default learning rate of 1e-2 and a weight decay of 5e-4 when not explicitly specified. GCN~\cite{kipf2017semi}, GAT~\cite{veličković2018graph}, and RGCN~\cite{zhu2019robust} have a fixed number of layers at 2.
For GCN and RGCN, the hidden dimension is chosen from $\{16, 32, 64, 128\}$. GAT's number of heads and head dimensions are selected from $\{1, 2, 4, 8, 16\}$ and $\{8, 16, 32, 64, 128\}$, respectively, with a total hidden dimension ranging from 16 to 128. Other baselines follow hyper-parameter combinations specified in their respective papers.
For DEGNN, $\alpha$ and $\beta$ are tuned from $\{0, 0.1, 1.0, 10\}$, the hidden dimension $D'$ is tuned from $\{128, 256, 512\}$. 
$k$ is searched in $\{1, 5, 10, 15, 20, 25\}$, $p$ and $q$ are searched in $\{0.2, 0.4, 0.6\}$.  The learning rate in the pre-training process is tuned from $\{$1e-2, 5e-3, 1e-3$\}$. 

\subsection{Semi-supervised Node Classification}
\subsubsection{Performance Comparison}
In this section, we evaluate the proposed DEGNN on semi-supervised node classification on original graphs.
The average accuracy and standard deviation of the model across 10 runs are summarized in Table~\ref{tab:node_classification}.
OOM indicates out of memory.
Based on the experiment, our proposed approach outperformed other models in citation networks (Cora, Citeseer), demonstrating the highest predictive accuracy.
This suggests that the experts successfully obtain superior representations for both edges and node features to eliminate latent noise within these graphs.
However, in co-purchasing networks (Photo, Computer), our model achieved competitive results but was marginally surpassed by the traditional GCN. Considering that other GSL models also lagged behind GCN in accuracy and the fact that GCN's predictive accuracy in co-purchasing networks is significantly higher than in citation networks, it is conceivable that the co-purchasing networks may contain less inherent noise within the original graphs, and the GSL models might be unnecessarily altering the graph.
\renewcommand{\arraystretch}{1.4}
\begin{table}
\caption{The results (accuracy(\%) $\pm$ std) of semi-supervised node classification on clean graphs. The top two performance is highlighted in bold and underline.}
\label{tab:node_classification}
\vspace*{-2mm}
\centering
\begin{adjustbox}{width=\textwidth} 
\begin{tabular}{c|ccc|ccc|cc}
\toprule
Dataset & GCN &  GAT &  RGCN & Pro-GNN & Rwl-GNN & STABLE & DEGNN-{\RomanI}  & DEGNN-{\RomanII} \\
\midrule
 Cora & $80.8\pm0.9$ & 79.5$\pm$0.8 & 79.6$\pm$0.7 & \underline{81.7$\pm$0.7} & 80.8$\pm$0.7 & 79.3$\pm$1.1 & \textbf{83.0$\pm$0.9}& 81.6$\pm$1.0\\
 Citeseer & 68.8$\pm$0.7 & 66.7$\pm$1.7 & 65.7$\pm$1.5 & 69.3$\pm$0.6 & 68.0$\pm$0.6 & 68.2$\pm$0.5 & \underline{70.9$\pm$1.9} & \textbf{71.6$\pm$1.4}\\
 Photo & \textbf{91.8$\pm$0.2} & 85.5$\pm$17.3 &91.7$\pm$0.5 & \underline{91.8$\pm$0.7} & 86.4$\pm$2.9 & OOM & 91.3$\pm$0.4 & 91.6$\pm$0.6\\
 Computers& \textbf{85.0$\pm$0.9} & 77.2$\pm$22.7 & 81.6$\pm$2.6 & OOM & 74.6$\pm$1.4 & OOM & \underline{83.5$\pm$0.9} & 82.8$\pm$0.8\\
\bottomrule
\end{tabular}
\end{adjustbox}
\end{table}
\subsubsection{Robustness Evaluation}
\begin{figure}[!t]
\includegraphics[width=\textwidth]{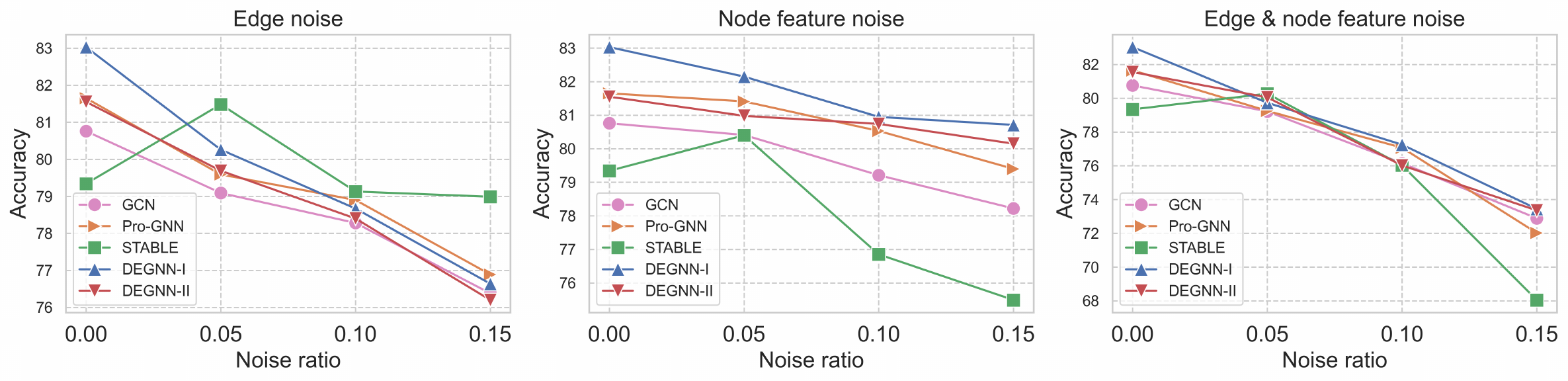}
\vspace*{-6mm}
\caption{Accuracy on Cora with noise added to either or both of edge and node features.}
\label{fig:cora_noise}
\end{figure}

In this experiment, we evaluate the robustness of the models by comparing their performance on graphs with noise added to either or both of the edge and node features.
The nature of these noise is described in Section~\ref{sec:noise}.
Specifically, we added edge noise with noise ratio 0.05, 0.1 and 0.15, while the parameter $\lambda$ for node feature perturbation is also set to 0.05, 0.1, and 0.15.
All experiments were conducted 10 times, and the average accuracy on Citeseer and Cora dataset are shown in Figures~\ref{fig:citeseer_noise} and Figure~\ref{fig:cora_noise}, respectively.
From this experiment, we can obtain the following observation:
\begin{figure}[!t]
\includegraphics[width=\textwidth]{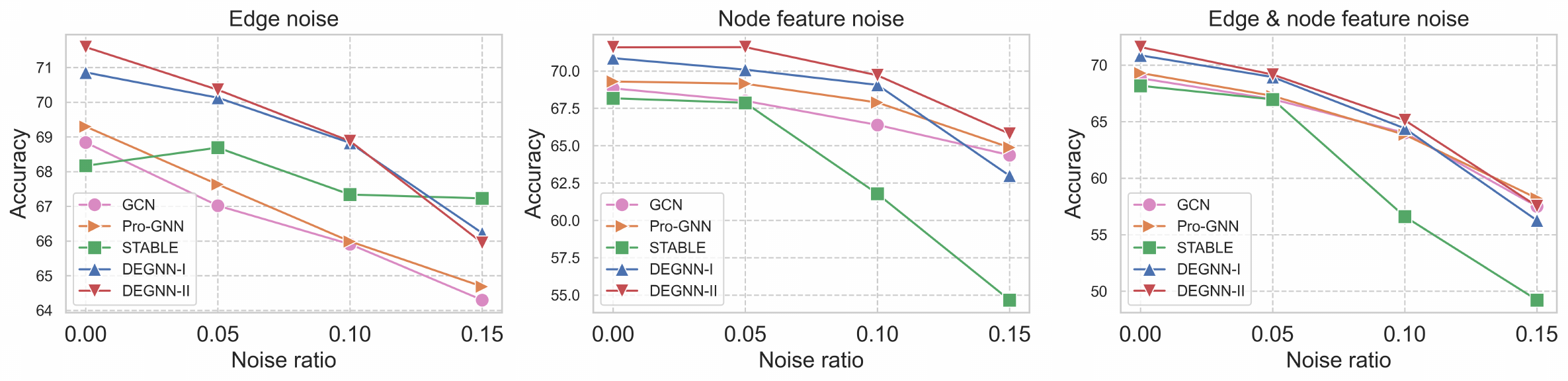}
\vspace*{-6mm}
\caption{Accuracy on Citeseer with noise added to either or both of edge and node features.}
\label{fig:citeseer_noise}
\end{figure}
\begin{itemize}
    \item When node feature noise is added, DEGNN-{\RomanI} or DEGNN-{\RomanII} demonstrated the highest accuracy among the compared models in all settings.
    \item When noise is introduced in only the edge and in both edge and node features, DEGNN demonstrated the best or competitive results compare to the baselines especially when the noise ratio is 0,  0.05 or 0.1.
    \item The smallest accuracy gap between edge noise ratio of 0 and 0.15 was observed in STABLE. However, it is highly vulnerable when node features are perturbed.
    \item GCN and Pro-GNN follow a similar trend, with Pro-GNN slightly outperforming GCN by a small margin. In many settings, their accuracy fell below that of DEGNN.
\end{itemize}

\subsection{Edge Expert Analysis}
To evaluate the edge expert, we compared the edge homophily ratio~\cite{zhu2020beyond} and node homophily ratio~\cite{Pei2020Geom-GCN} between the noisy graph and the graphs refined by the edge expert (referred to as the DEGNN graph) on various edge noise added settings.
In this experiment, we used DEGNN-{\RomanII}, and we set the hyper-parameter $k$, which controls the number of edge rewires, to 10\%.
\begin{figure}[!t]
\centering
\includegraphics[width=0.75\textwidth]{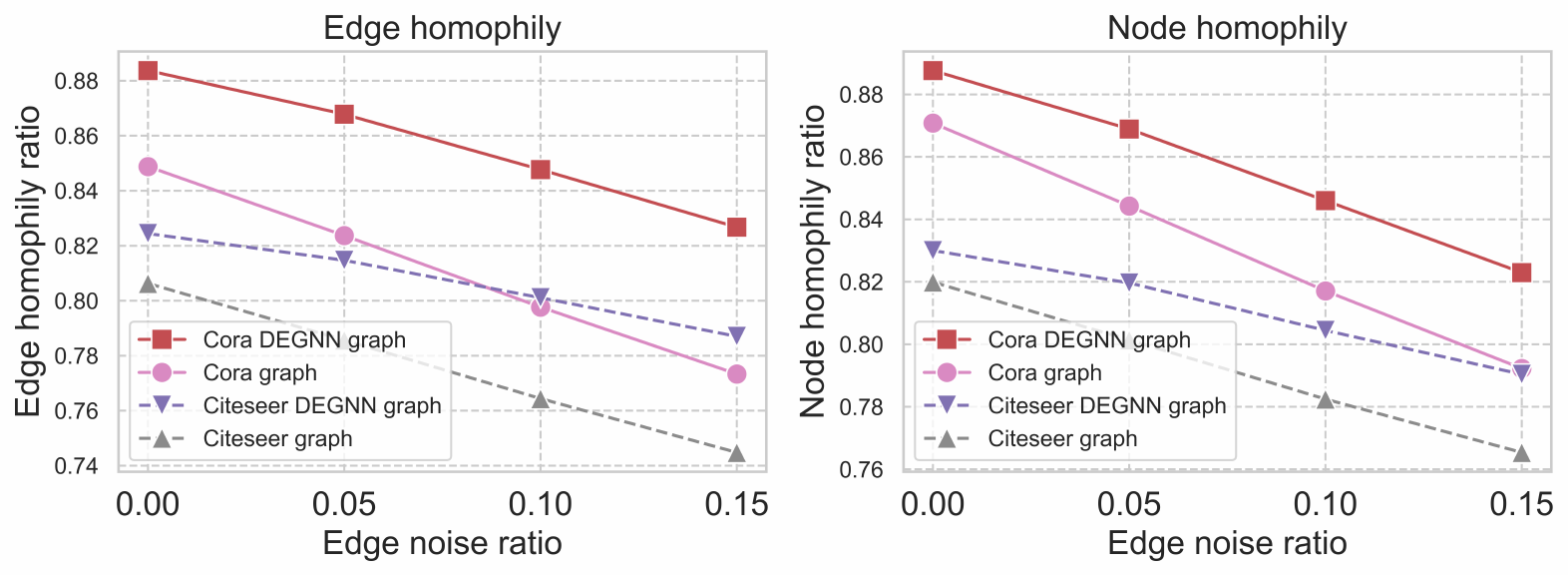}
\vspace*{-2mm}
\caption{Comparison of edge homophily ratio and node homophily ratio between noisy graphs and refined graphs by the edge expert on Cora and Citeseer.}
\label{fig:homophily}
\end{figure}
Figure~\ref{fig:homophily} shows the edge homophily ratio and node homophily ratio as edge noise is gradually added. The solid lines represent the results for the Cora dataset, while the dashed lines represent the results for the Citeseer dataset.
This experiment reveals that the edge expert consistently promotes high edge homophily and node homophily in all scenarios.


\subsection{Node Feature Expert Analysis}
\begin{figure}[!t]
\centering
\includegraphics[width=0.75\textwidth]{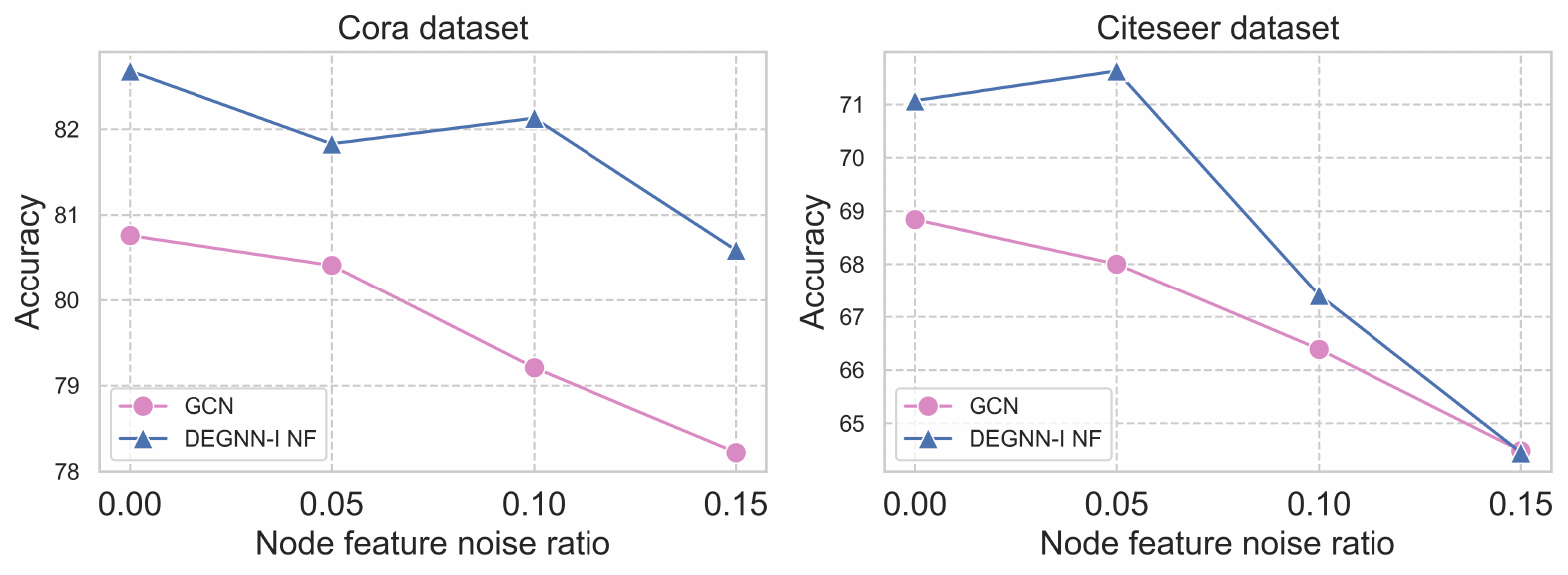}
\vspace*{-2mm}
\caption{Comparison of GCN and DEGNN-{\RomanI}  without edge expert on Cora and Citeseer with added node feature noise.}
\label{fig:nf_expert}
\end{figure}

In this experiment, we compared the effectiveness of the node feature expert by comparing the models' accuracy when node features have noise added.
Figure~\ref{fig:nf_expert} shows the average of accuracy of 10 runs of GCN and DEGNN-{\RomanI}  only having node feature expert (without edge expert) on Cora and Citeseer dataset with added node feature noise. Empirical results confirm that employing node embeddings derived from the node feature expert markedly elevates prediction accuracy across a majority of scenarios. This underscores both the indispensability and efficacy of the node feature expert.
\section{Conclusion}
In this paper, we identified the vulnerability of recent GSL methods to node feature noise and proposed a novel GNN model, DEGNN, to address this issue. 
DEGNN refines both edges and node features using two carefully designed experts via self-supervised learning, allowing it to robustly perform predictions in the presence of both edge and node feature noise. Extensive experiments verify the effectiveness of the experts in handling graphs with various noise scenarios.

In this paper, we employed GCN as both the encoder and downstream network.
Additionally, simple methods were used for augmentation and edge reconstruction techniques.
As future work, we plan to explore more specialized networks in the encoder and downstream network, as well as investigate different augmentation and edge reconstruction techniques.


\section*{Acknowledgements}
This work is partly supported by JSPS Grant-in-Aid for Scientific Research (grant number 23H03451, 21K12042) and the New Energy and Industrial Technology Development Organization (Grant Number JPNP20017).

%
%
%
%

\end{document}